\documentclass[10pt,twocolumn,letterpaper]{article}

\usepackage{iccv}
\usepackage{times}
\usepackage{epsfig}
\usepackage{graphicx}
\usepackage{amsmath}
\usepackage{amssymb}

\usepackage{multirow}
\usepackage{subfigure}
\usepackage[cmyk]{xcolor}
\usepackage{booktabs}
\usepackage{pifont}

\usepackage[pagebackref=true,breaklinks=true,colorlinks,bookmarks=false]{hyperref}

\iccvfinalcopy 

\begin{document}

\title{Coarse Is Better? A New Pipeline Towards Self-Supervised Learning with Uncurated Images}

\author{Ke Zhu, Yin-Yin He and Jianxin Wu \\
State Key Laboratory for Novel Software Technology \\
Nanjing University, Nanjing, China \\
{\texttt{\{zhuk,heyy\}@lamda.nju.edu.cn, wujx2001@nju.edu.cn}}
}

\maketitle

\begin{abstract}
    Most self-supervised learning (SSL) methods often work on curated datasets where the object-centric assumption holds. This assumption breaks down in uncurated images. Existing scene image SSL methods try to find the two views from original scene images that are well matched or dense, which is both complex and computationally heavy. This paper proposes a conceptually different pipeline: first find regions that are coarse objects (with adequate objectness), crop them out as pseudo object-centric images, then any SSL method can be directly applied as in a real object-centric dataset. That is, coarse crops benefits scene images SSL. A novel cropping strategy that produces coarse object box is proposed. The new pipeline and cropping strategy successfully learn quality features from uncurated datasets without ImageNet. Experiments show that our pipeline outperforms existing SSL methods (MoCo-v2, DenseCL and MAE) on classification, detection and segmentation tasks. We further conduct extensively ablations to verify that: 1) the pipeline do not rely on pretrained models; 2) the cropping strategy is better than existing object discovery methods; 3) our method is not sensitive to hyperparameters and data augmentations.
\end{abstract}

\section{Introduction}
\label{sec:intro}



Self-supervised learning (SSL) of visual representation has boosted the accuracy of various downstream tasks such as image classification~\cite{ImageNet} and object detection~\cite{faster-rcnn}. Recently, contrastive learning based SSL methods have become a popular paradigm. Representative methods in this family, such as MoCo~\cite{MOCOv2}, BYOL~\cite{BYOL} and SwAV~\cite{SwAV}, are based on the so-called \emph{object-centric} assumption: the images used for SSL are supposed to be object-centric, and hence two different \emph{views} augmented from the same image will share similar visual semantics. This assumption holds true in object-centric (also called curated) datasets like ImageNet~\cite{ImageNet}, in which objects are usually centered and occupy large areas. But, for downstream tasks like detection and segmentation, it simply breaks down. The images in such scene (or multi-object, or uncurated) datasets, \eg, VOC~\cite{VOC} and COCO~\cite{MS-COCO}, often contain many (small) objects per image. Hence, it is a natural requirement that we need new methods to handle SSL with scene images. In this paper, we aim to design a SSL method that \emph{works directly on scene images, \emph{without} relying on object-centric datasets like ImageNet}.
\begin{figure}
	\centering
	\includegraphics[width=0.9\linewidth]{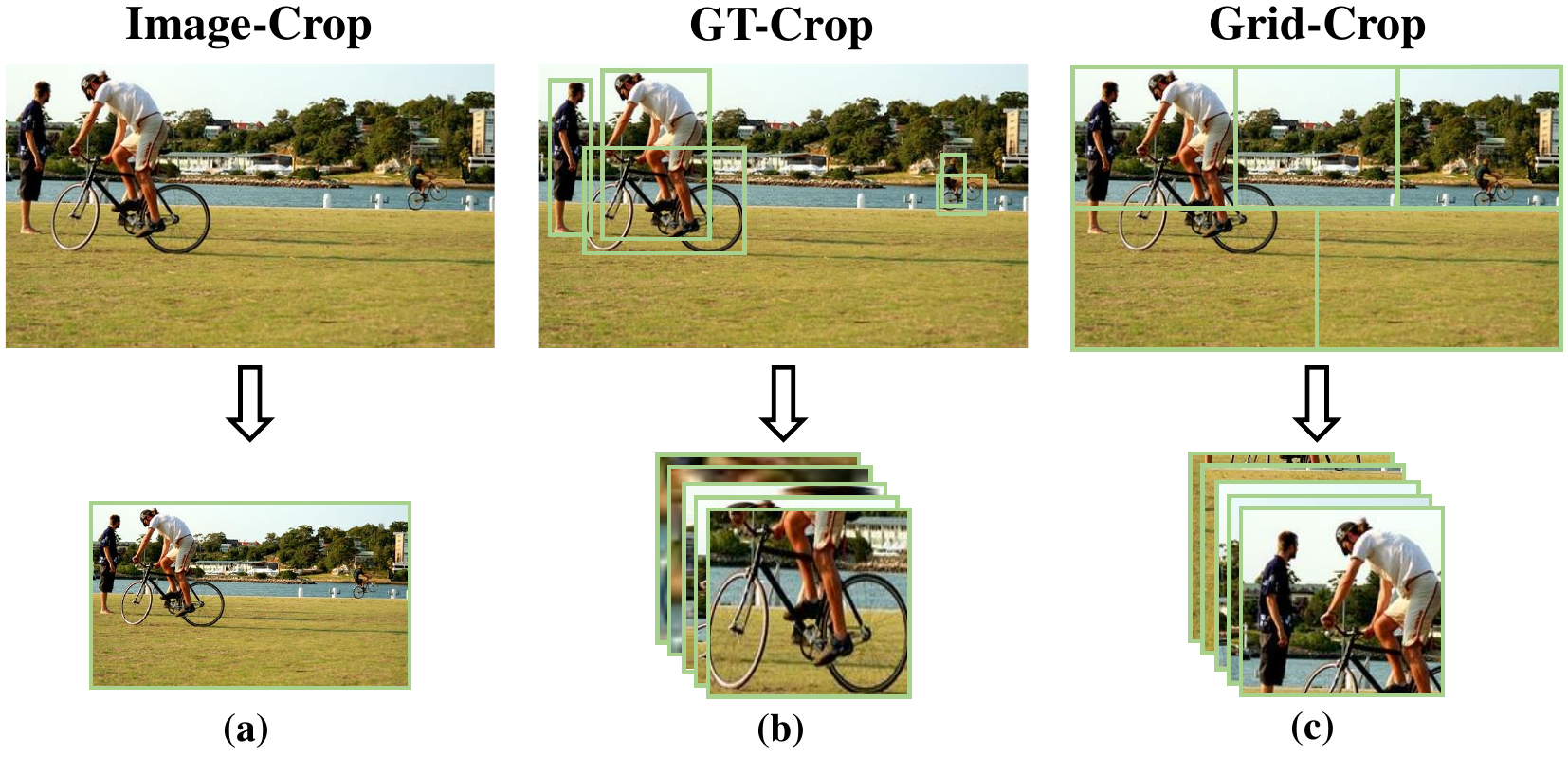}
	\caption{Three strategies for cropping pseudo object-centric images, using the entire image (a), groundtruth (b), and regular grid (c), respectively. This figure is best viewed in color.}
	\label{fig:crop}
\end{figure}
In SSL pretraining facing uncurated datasets, previous efforts are generally based on \emph{dense comparison}, such as DenseCL~\cite{DenseCL}, Self-EMD~\cite{self-EMD} and ReSim~\cite{ReSim}. They~\cite{self-EMD,DetCo,DetCon,Piont-Level-recent} introduce region-level or feature-level pretext tasks, with dedicated pipelines to formulate the loss functions. Another line of work~\cite{SoCo,Unsupervised_OBject} (\eg, SoCo~\cite{SoCo}) try to find pairs of precise boxes with similar semantics from images to apply the InfoNCE loss~\cite{InfoNCE}, but they involve heavy matching procedures~\cite{Unsupervised_OBject} on top of box generation (\eg, selective search~\cite{SS}). Besides, their downstream improvements are modest (\eg, considering SoCo pretrained on MS-COCO~\cite{MS-COCO}). 

One common characteristic of these prior arts is that they need \emph{manually} matched pairs on dense feature level or image level, and they \emph{generate the views directly from the raw input image}. The input image is uncurated, but the generated views are assumed to be good matches for each other and correspond to objects. This assumption is tremendously difficult to entertain. Besides, it is believed that the (unavailable) \emph{GT-box}/\emph{GT-mask} could lead to the best pretrained features (as shown in previous SSL method~\cite{DetCon}). The usefulness of coarse boxes (which contain partial objects) to SSL pretraining, however, remains unexplored.

In this paper, we propose a completely different \emph{scene image SSL paradigm without matching boxes}, and for the first time argue that \emph{coarse object boxes are comparable to GT boxes} for SSL pretraining. Our method can be summarized as follows: we first crop a few (\eg, 5) boxes from each uncurated image, \emph{only requiring the cropped boxes to coarsely contain objects} (\eg, do contain object(s), but not necessarily a single or a full object); then, the cropped boxes are treated as \emph{pseudo object-centric images}, which are directly fed to existing object-centric SSL methods (like MoCo-v2 or BYOL), and the two views are extracted from the same pseudo object-centric image for SSL learning.

\begin{table}
	\setlength{\tabcolsep}{4pt}
	\caption{Downstream tasks' results using different cropping strategies for object-centric SSL pretraining on VOC2007. All models (ResNet50) were pretrained for 800 epochs on the trainval set using BYOL or MoCo-v2. $AP_{50}^{bbox}$ evaluates object detection accuracy, and $mAP$ and $mAP^l$ are for multi-label classification with end-to-end finetuning and linear evaluation, respectively.}
	\label{table:motivation}
	\centering
	\footnotesize
	\begin{tabular}{lllll}
		\toprule[1pt]
		Method & Cropping  & AP$_{50}^{bbox}$ &  mAP & mAP$^{l}$\\
		\midrule[1pt]

		\multirow{4}{*}{BYOL} & Image-Crop & 63.3  & 63.0 & 39.8  \\
		& GT-Crop & 66.9 & 69.2 & 33.1\\
		& Grid-Crop & 68.1 & 70.3 & 41.6\\
		& Our-Crop & \textbf{69.5} & \textbf{70.7} & \textbf{42.5}\\
		\midrule
		\multirow{4}{*}{MoCo-v2} & Image-Crop & 61.8  & 62.2 & 26.9 \\
		& GT-Crop & 62.3 & 66.2 & 28.7\\
		& Grid-Crop & 65.6 & 67.7 & 37.2\\
		& Our-Crop & \textbf{67.2} & \textbf{69.6} & \textbf{44.9}\\
		\bottomrule[1pt]
	\end{tabular}
\end{table}

Our approach is motivated by the following simple experiment. We consider 3 different cropping strategies as in Fig.~\ref{fig:crop}. `Image-Crop' treats the entire input image as the only crop. `GT-Crop' utilizes the groundtruth bounding box annotations, and each bounding box becomes a crop. `Grid-Crop' splits the input image into 2 rows, with each row being split into 3 or 2 crops with equal size. These crops form pseudo object-centric images \emph{with dramatically different crop quality}, where `GT' is perfect but `Grid' is almost random in objectness. By pretraining an SSL model using these pseudo images to initialize downstream models, Table~\ref{table:motivation} shows their object detection and multi-label recognition accuracy on the VOC2007 dataset (more details in Sec.~\ref{sec:Method}). The results demonstrate that all cropping strategies (after SSL) are beneficial to downstream tasks. More importantly, the almost random `Grid' crop is even consistently better than the `GT' strategy. The `Grid' crops indeed contain objects, but the objects may be partial or multiple, i.e., only coarsely contain objects. Hence, we hypothesize that we do not need precisely cropped images as views, nor do we need matched crops. Instead, \emph{crops with coarse objects are what we want}. Treating these crops as pseudo object-centric images and then \emph{generate two views from the same pseudo image} is both much simpler and more beneficial.

We have then calculated the objectness of the crops, and consider two new crops: `Poor-Crop' (low objectness regions) and `GTpad-Crop' (GT boxes with padding). As shown in Table~\ref{table:crop-objectness} (cf. the visualization in Fig.~\ref{fig:objness-mAP}), too low objectness (`Poor-Crop') is detrimental for SSL, while too high objectness (`GT-Crop') leads to suboptimal results compared with medium objectness (`GTpad-Crop' and `Our-Crop'). These results again support our hypothesis: \emph{objectness should not be too high, and coarse crops are enough}. On the other hand, \emph{a filtering algorithm is required to filter away crops with low objectness}. For example, roughly 40\% of the crops in `Grid-Crop' have lower than 20\% objectness (see appendix for more details), which hinders the model from learning quality representation.

\begin{table}
	\setlength{\tabcolsep}{4pt}
	\caption{Empirical analysis of cropping strategies. `Objectness' is the average ratio of object pixels within each crop. We selected a best padding ratio for `GTpad-Crop' and generated `Poor-Crop' objectness around 20\%. Models were pretrained with BYOL. We used the pixel-level groundtruth annotations only in calculating the objectness.}
	\label{table:crop-objectness}
	\centering
	\footnotesize
	\begin{tabular}{lllll}
		\toprule[1pt]
		  Cropping  & AP$_{50}^{bbox}$ &  mAP & mAP$^{l}$ & Objectness \\
		\midrule[1pt]
		 GT-Crop & 66.9 & 69.2 & 33.1 & 100.0\% \\
		 GTpad-Crop & \textbf{69.7} & \textbf{72.3} & \textbf{40.2} & 48.2\% \\
		 \midrule
	     Poor-Crop & 60.4 & 62.2 & 21.4 & 20.1\%\\
         Image-Crop & 63.3  & 63.0 & 39.8 & 36.4\%\\
         Grid-Crop & 68.1 & 70.3 & 41.6 & 39.8\% \\

		 Our-Crop & \textbf{69.5} & \textbf{70.7} & \textbf{42.5} & 52.6\%\\
		\bottomrule[1pt]
	\end{tabular}
\end{table}

Consequently, we propose a novel crop generation and filtering method to obtain bounding boxes with adequate objectness, and demonstrate its superiority over existing unsupervised object discovery methods from both qualitative and quantitative results. The effectiveness of our new pipeline and the cropping strategy are extensively verified on uncurated datasets like MS-COCO and VOC for detection, segmentation and classification, and are further studied by abundant ablations. Our contributions can be summarized as:

\begin{enumerate}
  \item We conduct an empirical study to argue that cropping benefits scene images SSL---just treat these local crops as pseudo object-centric images.
  \item We verify that we do not need precisely matched boxes for scene SSL pretraining, and propose a novel box generation and filtering strategy to obtain crops with coarse objectness (\ie, `Our-Crop'). 
  \item Feeding the cropped pseudo images to existing SSL methods (\eg, BYOL, DenseCL or MAE), we achieve consistent improvements on VOC and MS-COCO for recognition, detection and segmentation. Note that we do \emph{not} rely on the ImageNet pretraining data.
\end{enumerate}

\section{Related Work}


\textbf{SSL on object-centric images.} Self-supervised learning has become a powerful tool to learn good representations from unlabeled data~\cite{PR_Jigsaw_ViT,PR_LPCL}. In particular, contrastive SSL methods~\cite{Coding-contrastive,PR_GraphSSL,prototypical,PR_faceSSL} become the most popular and have largely boosted the performance of various downstream tasks. Previous works formulated the InfoNCE~\cite{InfoNCE} loss using contrastive predictive coding, and InfoMin~\cite{InfoMin} argued that the mutual information of two views should be reduced as long as the task-relevant information are kept intact. Recent studies based on bootstrapping~\cite{BYOL,PixPro}, memory bank~\cite{MOCOv2,DetCo,Piont-Level-recent,DetCon} and online learning~\cite{SwAV} tried to pull positive pairs closer while pushing negative pairs apart. The underline assumption of these methods is that two positive views generated from the same object-centric image should share similar content, which may cease to be true in uncurated datasets. \cite{SSL_in_the_wild} attempted to apply SSL in the wild, and~\cite{NIPs_revisiting} tried to improve SSL of scene images by introducing constrained multi-crop~\cite{SwAV}. Still, a specialized method for uncurated scene images is of high importance. 

\textbf{SSL on uncurated scene images.} Recent works~\cite{DenseCL,MaskCo,self-EMD,MLS} started to focus on the usage of uncurated datasets like MS-COCO~\cite{MS-COCO}, aiming to get better representation for dense prediction tasks. DenseCL~\cite{DenseCL}, MaskCo~\cite{MaskCo}, Self-EMD~\cite{self-EMD} tried to form positive pairs with dense features. But, they need manually matched pairs from feature level or image level, which can be difficult to entertain, especially in scene images with multiple different objects. 

Another SSL paradigm~\cite{Unsupervised_OBject,SoCo,DetCon} resorted to external unsupervised object discovery methods such as selective search~\cite{SS} to obtain precise object boxes in scene images. They tried to learn a good representation based on the global image and local object parts. These methods usually involved complex filtering algorithms like K-Nearest Neighbors~\cite{Unsupervised_OBject}, so as to select object bounding boxes as precisely as possible. They assumed that the more precise the object proposals, the better the SSL. In this paper, however, we call into question the validity of this assumption. Instead, we argue that bounding boxes that are only roughly objects (with adequate objectness) are enough, and we need to crop them as pseudo object-centric images. Then, we advocate that the two positive views need to be cropped within the same one such pseudo image, \emph{not from two original input images}.

A recent work that is close to ours is~\cite{objaware}, which try to obtain views from object-aware boxes. However, we clarify that Our-Crop and ~\cite{objaware} is much different. First,~\cite{objaware} directly rely on existing object discovery methods (some even need groundtruth annotations), while our cropping strategy is done in a fully unsupervised manner. More importantly, \emph{quantitative analysis} of how different quality (objectness, cf. Table 1-2) of boxes affect pretraining are not told in their paper, while we found perfect GT are \emph{not} possibly the best and argued \emph{coarse} is enough. Technically, they are very different, too: they add extra augment like dilation, shifting to the BING boxes, adjust many augmentation parameters, and adopt different projection heads, while we \emph{directly} fed our pseudo-images into SSL without \emph{any} modification. In a word, our pipeline, cropping strategy and the analysis shares minimal similarity with that of~\cite{objaware}.

\section{Method}
\label{sec:Method}

We first briefly introduce the formulation of two SSL methods for object-centric images, as they are both our baseline and component techniques. Then, we present the overall pipeline of the proposed method in detail. Finally, we present the key component of our pipeline: a novel cropping strategy to generate pseudo object-centric images.

\subsection{MoCo-v2 and BYOL}
\label{sec:formultion}

Both MoCo-v2 and BYOL implicitly follow the object-centric assumption, but in practice we can feed any type of images (\eg, uncurated ones) to them. MoCo-v2~\cite{MOCOv2} is based on the InfoNCE~\cite{InfoNCE} loss:
\begin{equation}
\mathcal{L}_q = - \log \frac{\exp{ (q\cdot k_+/\tau)}}
 {\exp{(q \cdot k_+/\tau)} + \sum_{k_-} \exp{(q \cdot k_- / \tau)}} \,,
\end{equation}
where $\tau$ is the temperature, and $q$ is the encoded query that computes the loss over one positive key $k_+$ and many negatives $k_-$. Note that $q$ and $k_+$ are \emph{two views generated from the same image by different data augmentations}.

BYOL~\cite{BYOL} devises an online network $\theta$ and a slow moving averaged target network $\xi$, which was inspired by several earlier reinforcement learning methods~\cite{BYOL_prior}. The loss function of BYOL is designed to only pull positive views together:
\begin{equation}
 \mathcal{L}_{\theta, \xi} = \left\| \frac{q_{\theta}}{\|q_{\theta}\|} - \frac{z'_{\xi}}{\|z'_{\xi}\|} \right\|_2^2 
 = 2 \left( 1 - \frac {\langle q_{\theta}, z'_{\xi} \rangle} {||q_{\theta}||\cdot ||z_{\xi}'||} \right) \,,
\end{equation}
in which $q_{\theta}$ and $z'_{\xi}$ are feature representations for the online and the target network, respectively. Note that $q_{\theta}$ and $z'_{\xi}$ are also \emph{two views generated from the same image}. Many recent work~\cite{self-EMD,Unsupervised_OBject,MaskCo} have followed BYOL for its efficacy.

\subsection{A new pipeline for SSL with scene images}
\label{sec:pipeline}

It is easy to accept that two views generated out of the same object-centric image share enough semantic similarity. But, as suggested by previous work~\cite{DenseCL,MaskCo,Unsupervised_OBject}, two views generated from one non-object-centric image potentially have insufficient semantic overlap and will lead to suboptimal accuracy in downstream tasks. Hence, we argue that instead of following the commonly used strategy to find semantically or visually matched object regions as the two views, it is better to \emph{crop out regions that coarsely contain objects (\ie, with adequate objectness) as pseudo object-centric images}, then \emph{generate two views from the same one such pseudo image}. The two views generated in this manner are naturally similar to each other, and will contain coarse object information, too.

\begin{figure*}
	\centering
	\includegraphics[width=0.8\linewidth]{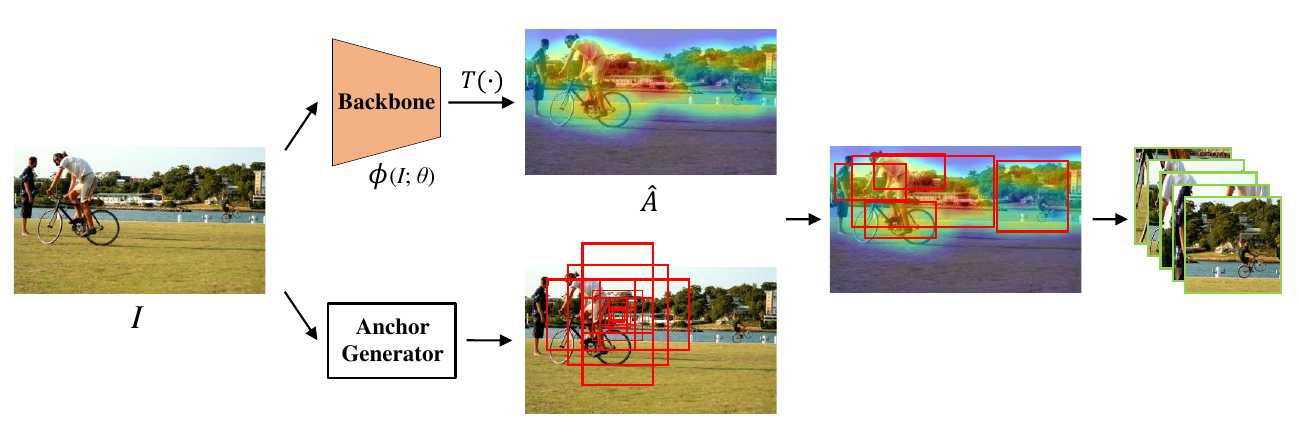}
	\caption{The proposed cropping strategy. In the top branch, an image $I\in\mathbb{R}^{H\times W\times C}$ first passes through a self-supervised pretrained backbone $\phi(\cdot)$ (\eg, ResNet-50) to obtain a feature tensor $\boldsymbol{x}\in \mathbb{R}^{h\times w \times d}$, which is then transformed by $T(\cdot)$ (sum, normalize and resize) to an objectness score map $\hat{A}\in \mathbb{R}^{H\times W}$ (color coded on top of the input image). Second, in the bottom branch, a large number of anchor boxes with different sizes and aspect ratios are generated. Third, anchors are filtered according to the score map $\hat{A}$. Finally, the top-5 anchors are our selected crops (pseudo object-centric images). Best viewed in color.}
	\label{fig:main-method}
\end{figure*}

The proposed pipeline is as follows:
\begin{enumerate}
	\item Given a scene image dataset like COCO, use any object-centric SSL methods to learn a backbone model (the $\phi(\cdot)$ shown in Fig.~\ref{fig:main-method}) by simply pretending these images as object-centric ones (though Table~\ref{table:COCO-FPN-stages} proved that this backbone learning stage is \emph{not} necessary).
	\item For each uncurated image, use our proposed cropping strategy and this backbone model $\phi(\cdot)$ to find a few (\eg, 5) crops that coarsely contain objects: filter away crops with low objectness and the remaining ones contain adequate objectness. These crops are treated as pseudo object-centric images.
	\item Feed these pseudo object-centric images to any object-centric SSL method to train a randomly initialized backbone model from scratch (that is, we do \emph{not} inherit the weights of $\phi(\cdot)$ from stage 1) so as to finish unsupervised visual representation learning with uncurated images.
\end{enumerate}

Our pipeline (and its feasibility) is motivated by the experiments in Table~\ref{table:motivation}, in which we tried 3 cropping strategies and 2 SSL methods (BYOL and MoCo-v2) on VOC2007. Following~\cite{Tobias}, we set batch size to 256, and trained 800 epochs on VOC2007 trainval~\cite{VOC}. For the downstream tasks, we trained Faster R-CNN~\cite{faster-rcnn} for object detection and finetuned multi-label classification models. 

As shown in Table~\ref{table:motivation}, directly cropping two views from the uncurated images (`Image-Crop') works fairly poorly, since it has a big chance to generate two views containing different semantics. `GT-Crop' is much better, showing the importance of cropping two views from objects locally. What is surprising is that cropping two views from one fixed-grid pseudo image (`Grid-Crop') achieves even higher mAP than `GT'! In fact, the five pseudo images of `Grid' contain random contents and are at best coarse objects, which suggests that we do \emph{not need precise object locations} in a cropping strategy. Then, it is natural to hypothesize that \emph{constructing two views within one locally cropped image with coarse object(s) is the key}---we do not need precise object cropping, and it is both easier and better than finding matching regions as the views~\cite{DenseCL,DetCon,Unsupervised_OBject} either from the image-level or the feature-level.

Meanwhile, as Table~\ref{table:crop-objectness} suggests, the pseudo images in `Grid' are too coarse such that a large proportion of them have extremely low quality (about 40\% of the crops in `Grid' have lower than 20\% objectness), and may contain no objects at all (\eg, the one contains only grass in Fig.~\ref{fig:crop}). The diversity of the boxes is not guaranteed, either. These drawbacks necessitate a novel strategy that filters away cropped pseudo images with low objectness scores and to make the cropped pseudo images diverse (\eg, it is harmful if all 5 crops are around one same region).

\subsection{Generating coarse object crops}
\label{sec:framework}

The proposed cropping strategy (Fig.~\ref{fig:main-method}) differs from previous unsupervised object discovery methods~\cite{SS,BING,EdgeBox}, since these object discovery methods focus on cropping highly accurate objects (as later verified in Fig.~\ref{fig:other_crop_boxtype_ratio}), which may not suit SSL pretraining well (as indicated by Table~\ref{table:crop-objectness} and compared in Table~\ref{table:other-crop}). Instead, we argue that crops that coarsely contain objects are enough. As Fig.~\ref{fig:main-method} illustrates, we compute both the per-pixel objectness score and a set of predefined anchors, then filter the anchors with the objectness scores.

\textbf{Anchor generation.} We borrow the anchors from object detection~\cite{faster-rcnn,Mask-RCNN}, since it \emph{does not} need a real forward pass and is highly efficient. Besides, the diversity of the anchors are guaranteed, as shown by the detection literature. For an image $I$ with height $H$ and width $W$, a feature tensor $\boldsymbol{x}\in \mathbb{R}^{d\times h \times w}$ would be obtained \emph{if it were} to be sent to the backbone network, where $d$, $w$ and $h$ are the dimensionality, height and width of the feature tensor, respectively. For each spatial location out of the $h\times w$ in the feature tensor, we predefine 12 anchors with four different sizes (32, 64, 128, 256) and three aspect ratios (0.5, 1.0, 2.0). Then the spatial coordinates of these anchors are mapped from the feature space back into the image space according to the down-sampling ratio $H/h$. Thus, a total of $12 \times h \times w$ anchor boxes are generated for an image. 

\textbf{Objectness scores and crops.} The key idea to filter the generated boxes is to assign each box an objectness score. For simplicity, we generate a shared per-pixel score map, and the objectness score of a box is the average score of all pixels within it. Following SCDA~\cite{SCDA}, an image $I\in \mathbb{R}^{H\times W}$ is first sent to a feature extractor $\phi$ (the backbone network) to obtain the feature tensor $\boldsymbol{x} \in \mathbb{R}^{d\times h \times w}$: $ \boldsymbol{x} = \phi(I;\theta)$, where $\theta$ are the backbone's parameters. Then, a sum operation is done on the feature dimension to obtain a corresponding score map $A\in \mathbb{R}^{h\times w}$:
\begin{equation}
    A_{i,j} = \sum_d x_{i,j,d} \,,
\end{equation}
in which $i\in \{1,2,...,h\}$ and $j\in \{1,2,...,w\}$. As pointed out by SCDA~\cite{SCDA}, the score map $A$ is indicative of the objectness at the corresponding position. We then linearly normalize the score map $A$ to the $[0,1]$ range and upsample it to the original image size by bilinear interpolation to get the final score map $\hat{A}\in \mathbb{R}^{H\times W}$. 

Suppose we have generated $N$ anchors in an image, namely, $\boldsymbol{c}_1, \boldsymbol{c}_2, \dots, \boldsymbol{c}_N$. We calculate the mean score within each anchor by using the final score map $\hat{A}$. Specifically, for a given anchor $\boldsymbol{c}_k$, the score $s_{k}$ for it is:
\begin{align}
    s_{k} = \frac{1}{h' \times w'}\sum_{i=x_1}^{y_1} \sum_{j=y_1}^{y_2} \hat{A}_{i,j} \,,
\end{align}
where $h'$ and $w'$ are the height and width of the anchor $\boldsymbol{c}_k$, and $(x_1,y_1)$ and $(x_2,y_2)$ are the top-left and bottom-right coordinates of the anchor $\boldsymbol{c}_{k}$, respectively. The NMS~\cite{NMS} operation is then applied to merge heavily-overlapped anchors. Finally, anchors with the top-5 scores are cropped.

\textbf{The backbone and the final learning.} This box filtering process needs a feature extractor $\phi$ to produce an objectness score map $A$. Take Table~\ref{table:motivation} for example, we use MoCo-v2 and uncurated images (i.e., `Image-Crop') to learn the backbone $\phi$. After all the pseudo object-centric images are cropped, we \emph{randomly initialize} a model, and then perform SSL using any object-centric SSL method and these pseudo images. As shown in Table~\ref{table:motivation}, `Our-Crop' significantly outperforms other cropping strategies, showing the effectiveness of our cropping strategy. 

\section{Experiments}
\label{sec:exp}
For almost all of our experiments, we choose MS-COCO~\cite{MS-COCO} as the pretraining dataset and evaluate the effectiveness of our cropping strategy on various downstream benchmarks: MS-COCO detection and segmentation, VOC0712~\cite{VOC} detection, CityScapes~\cite{Cityscapes} segmentation and 7 small classification~\cite{cub200} datasets. We also explore the effectiveness of Our-Crop pretrained on truly object-centric dataset (\eg, ImageNet~\cite{ImageNet} only in Table~\ref{table:imagenet-results}).

\subsection{Experimental settings}

\textbf{Datasets.} MS-COCO is a large uncurated dataset. We use 118k training images to pretrain our backbone with different self-supervised methods (MoCo-v2, BYOL, DenseCL and MAE). VOC2007 contains 5,011 trainval and 4,952 test images, while VOC2012~\cite{VOC} has 11k trainval images. The Cityscapes dataset is a scene dataset containing 3475 images (2975 for training and 500 for validation). Since most previous~\cite{DenseCL,NIPs_revisiting} SSL works utilize these datasets for evaluation, we evaluate the object detection and instance segmentation tasks on MS-COCO, object detection and multi-label classification tasks on VOC, and semantic segmentation on Cityscapes. Besides, we also validate our effectiveness on various single-label classification datasets, namely CUB200~\cite{cub200}, Flowers~\cite{flowers}, Cars~\cite{cars}, Aircraft~\cite{aircrafts} and DTD~\cite{DTD}. On top of that, we explore ImageNet~\cite{ImageNet} (128k train images) pretraining to evaluate the generalization ability of our cropping strategy.

\textbf{Training details.} We pretrain all models from scratch for 400 epochs using ResNet50~\cite{ResNet} (except ViT-B~\cite{ViT} for MAE~\cite{MAE}) as our backbone with synchronized batch normalization~\cite{syncBN} during pretraining. Specifically, for MoCo-v2~\cite{MOCOv2}, batch size and learning rate are 256 and 0.3, respectively. The temperature $\tau$ is 0.2, and MLP's hidden dimension is 2048. For data augmentation, we follow the default setting of~\cite{MOCOv2}. For BYOL, we set the batch size as 512 and follow the base learning rate scheduler of~\cite{PixPro,SoCo}. The momentum grows from 0.99 to 1 using the cosine scheduler, and data augmentation follows the original paper~\cite{BYOL}. DenseCL~\cite{DenseCL} is suitable for uncurated images, which follows most of the settings from MoCo-v2 except for a different (dense) projection head. For MAE, we strictly follow its original pre-training settings~\cite{MAE}.

For simplicity, we use MoCo-v2 pretrained with `Image-Crop' as the backbone model $\phi(\cdot)$ to generate our crops (5 per image, cf. Fig.~\ref{fig:main-method} for a full illustration) on the MS-COCO dataset. We then train MoCo-v2, BYOL, DenseCL and MAE \emph{from scratch} by generating two views from one of these crops (pseudo images), referred to as `MoCo-v2$^{our}$', `BYOL$^{our}$', `DenseCL$^{our}$' or `MAE$^{our}$', respectively. 

\textbf{Downstream finetuning.} After the MoCo-v2, BYOL, DenseCL or MAE model is pretrained on our cropped pseudo object-centric images, we finetune them for the MS-COCO object detection or instance segmentation, VOC detection and Cityscapes segmentation tasks. Specifically, we finetune for 90k iterations on MS-COCO using Mask R-CNN R50-FPN and Mask R-CNN R50-C4, and finetune for 24k iterations on the VOC07+12 trainval set using Faster R-CNN R50-C4. On Cityscapes, we finetune for 40k iterations using PSANet. The learning rate and batch size are 0.02 and 16, respectively. We strictly follow the settings in ViTDet~\cite{ViTDet} for MAE pretrained models. Following previous works~\cite{BYOL,PSPNet}, we adopt AP (average precision) for detection and instance segmentation, and adopt mAP (mean average precision) for multi-label classification. The mIoU (mean IoU), mAcc (mean accuracy) and aAcc are used for semantic segmentation. We take 3 runs on VOC detection and CityScapes segmentation tasks since these results have larger variances. All our experiments were conducted using PyTorch~\cite{pytorch} and we used 8 GeForce RTX 3090 for our experiments. More details can be found in the appendix.

\subsection{COCO detection and segmentation.}
\begin{table}
	\setlength{\tabcolsep}{1.6pt}
	\footnotesize
	\centering
	\caption{Downstream results of COCO detection and segmentation using the Mask R-CNN R50-FPN structure. Models were pretrained 400 epochs from scratch using BYOL, MoCo-v2 and DenseCL on MS-COCO with cropped pseudo images (`$our$', 5 crops per image by default). `IN' and `CC' stand for ImageNet and COCO, respectively. The SSL method SoCo$^*$ is our reproduced result.}
	\label{table:COCO-FPN}
	\begin{tabular}{llllllll}
		\toprule[1pt]
		\multirow{2}{*}{Method} & \multirow{2}{*}{Data}  & \multicolumn{3}{c}{Detection} & \multicolumn{3}{c}{Segmentation}  \\

		& & AP$^{bbox}$ & AP$^{bbox}_{50}$ & AP$^{bbox}_{75}$ & AP$^{seg}$ & AP$^{seg}_{50}$ & AP$^{seg}_{75}$\\
		\midrule[1pt]
		 \textcolor{lightgray}{Supervised} &  \textcolor{lightgray}{IN} & \textcolor{lightgray}{38.9} &  \textcolor{lightgray}{59.6} &  \textcolor{lightgray}{42.7} &  \textcolor{lightgray}{35.4} &  \textcolor{lightgray}{56.5} &  \textcolor{lightgray}{38.1}   \\
		\midrule
		ReSim-C4~\cite{ReSim} & IN & 39.3 & 59.7 & 43.1 & 35.7 & 56.7 & 38.1 \\
		LEWEL$_{M}$~\cite{LEWEL} & IN & 40.0 & 59.8 & 43.7 & 36.1 & 57.0 & 38.7 \\
		SimCLR~\cite{SimCLR}   & CC & 37.0 & 56.8 & 40.3 & 33.7 & 53.8 & 36.1    \\
		Self-EMD~\cite{self-EMD} & CC & 39.3 & 60.1 & 42.8 & - & - & - \\
		SoCo$^*$~\cite{SoCo} & CC & 39.1 & 59.1 & 42.7 & 35.4 & 56.0 & 37.8 \\
            \midrule
		MoCo-v2 & CC &  38.2 & 58.0 & 41.9 & 34.7 & 55.1 & 37.2\\
		MoCo-v2$^{our}$ & CC &  \textbf{39.3} & \textbf{59.1} & \textbf{42.7} & \textbf{35.6} & \textbf{56.4} & \textbf{38.0}\\
		\midrule
		DenseCL & CC & 39.3 & 58.9 & 42.9 & 35.4 & 56.0 & 37.8 \\
		DenseCL$^{our}$ & CC & \textbf{39.9} & \textbf{59.8} & \textbf{43.5} & \textbf{35.9} & \textbf{56.8} & \textbf{38.5} \\
		\midrule
		BYOL & CC  & 38.8 & 58.5 & 42.2 & 35.0 & 55.9 & 38.1 \\
		BYOL$^{our}$  & CC &  \textbf{40.2} & \textbf{60.4} & \textbf{43.9} & \textbf{36.4} & \textbf{57.3} & \textbf{39.0}\\
		\bottomrule[1pt]
	\end{tabular}
\end{table}
\textbf{ResNet models.} We first evaluate on MS-COCO object detection and instance segmentation using ResNet-50 as backbone. The results of MoCo-v2 and BYOL pretrained models using the Mask R-CNN R50-FPN structure are in Table~\ref{table:COCO-FPN}. Specifically, our method has 1.4\%, 1.1\% and 0.6\% AP$^{bbox}$ gains over the BYOL, MoCo-v2, and DenseCL baseline, respectively. Our method also achieves state-of-the-art performance within self-supervised training methods on uncurated images (i.e., ImageNet free). It outperforms ReSim~\cite{ReSim} and SoCo~\cite{SoCo} consistently. It also surpasses its supervised counterpart by 1.3\% AP$^{bbox}$ and 1.0\% AP$^{seg}$ for detection and segmentation, respectively. We also tested our method using the Mask R-CNN R50-C4 structure. As shown in Table~\ref{table:COCO-C4}, our cropping strategy has roughly 1\% AP$^{bbox}$ gain consistently when compared with the baseline MoCo-v2, BYOL and DenseCL (i.e., `Image-Crop'). More importantly, DenseCL with our cropping strategy has surpassed supervised pretraining by a large margin (1.2\% AP$^{bbox}$ and 1.1\% AP$^{seg}$ for detection and segmentation, respectively).

\begin{table}
	\centering
	\footnotesize
	\setlength{\tabcolsep}{1.6pt}
	\caption{Downstream results on COCO with Mask R-CNN R50-C4. All models were pretrained on MS-COCO for 400 epochs.}
	\label{table:COCO-C4}
	\begin{tabular}{llllllll}
		\toprule[1pt]
		\multirow{2}{*}{Method} & \multirow{2}{*}{Data}  & \multicolumn{3}{c}{Detection} & \multicolumn{3}{c}{Segmentation}  \\

		& & AP$^{bbox}$ & AP$^{bbox}_{50}$ & AP$^{bbox}_{75}$ & AP$^{seg}$ & AP$^{seg}_{50}$ & AP$^{seg}_{75}$\\

		\midrule[1pt]
		 \textcolor{lightgray}{Supervised} &  \textcolor{lightgray}{IN} &  \textcolor{lightgray}{38.1} &  \textcolor{lightgray}{58.1} &  \textcolor{lightgray}{41.1} &  \textcolor{lightgray}{33.2} &  \textcolor{lightgray}{54.8} &  \textcolor{lightgray}{35.0}\\

		\midrule
		BYOL  & CC & 36.9 & 56.7 & 39.4 & 32.4 & 53.5 & 34.3    \\

		BYOL$^{our}$ & CC &  \textbf{37.9} & \textbf{57.8} & \textbf{40.7} & \textbf{33.1} & \textbf{54.3} & \textbf{35.2}\\
		\midrule
		MoCo-v2   & CC & 37.3 & 56.7 & 40.4 & 32.8 & 53.5 & 34.9    \\

		MoCo-v2$^{our}$ & CC &  \textbf{38.1} & \textbf{57.4} & \textbf{41.4} & \textbf{33.4} & \textbf{54.2} & \textbf{35.8}\\
		\midrule
		DenseCL & CC & 38.3 & 57.9 & 41.4 & 33.5 & 54.4 & 35.7 \\
		DenseCL$^{our}$ & CC & \textbf{39.3} & \textbf{58.8} & \textbf{42.4} & \textbf{34.3} & \textbf{55.6} & \textbf{36.6} \\
		\bottomrule[1pt]
	\end{tabular}
\end{table}

\textbf{ViT structure with MAE.} Since our cropping strategy is orthogonal to SSL methods, we now validate its adaptability to the non-contrastive SSL method MAE~\cite{MAE}. We pretrain ViT-B on COCO with original image and Our-Crop (`our') using MAE for 800 epochs and finetune COCO detection. Our cropping strategy surpasses the original Image-Crop by a significant margin for both Mask RCNN and Cascade RCNN~\cite{Cascade} detectors, as Table~\ref{table:MAE} shows. We argue that decoupling original image to a set of image patches (Our-Crop) could \emph{relieve the difficulty of the MAE's reconstruction}, since the pixels in one of these pseudo object-centric images will have higher correlation to each other that jointly describe a certain object content. 

\begin{table}
	\centering
	\footnotesize
	\setlength{\tabcolsep}{1.8pt}
	\caption{ViT-Base pretrained (800ep) on COCO with MAE~\cite{MAE}, and finetuned on COCO using Mask RCNN FPN (`Mask') and Cascade RCNN FPN (`Cas.') according to ViTDet~\cite{ViTDet}.}
	\label{table:MAE}
	\begin{tabular}{llllllll}
		\toprule[1pt]
		\multirow{2}{*}{Method} & \multirow{2}{*}{Detector}  & \multicolumn{3}{c}{Detection} & \multicolumn{3}{c}{Segmentation}  \\

		& & AP$^{bbox}$ & AP$^{bbox}_{50}$ & AP$^{bbox}_{75}$ & AP$^{seg}$ & AP$^{seg}_{50}$ & AP$^{seg}_{75}$\\
		\midrule[1pt]
		\textcolor{lightgray}{Random} &  \textcolor{lightgray}{Mask} &  \textcolor{lightgray}{27.8} & \textcolor{lightgray}{45.4} & \textcolor{lightgray}{29.4} & \textcolor{lightgray}{26.0} & \textcolor{lightgray}{43.0} & \textcolor{lightgray}{27.0}\\

		MAE & Mask &  38.0 & 57.8 & 41.1 & 34.6 & 55.0 & 37.0\\
		MAE$^{our}$  &Mask & \textbf{43.2} & \textbf{63.8} & \textbf{47.5}  & \textbf{39.0} & \textbf{60.8} & \textbf{41.8}    \\
		\midrule
		\textcolor{lightgray}{Random} &\textcolor{lightgray}{Cas.} & \textcolor{lightgray}{31.9} & \textcolor{lightgray}{47.0} & \textcolor{lightgray}{34.1} & \textcolor{lightgray}{28.0} & \textcolor{lightgray}{44.7} & \textcolor{lightgray}{29.8}    \\
		MAE & Cas.& 41.7 &  58.8 & 45.1 & 36.3 & 56.2 & 39.1 \\
		MAE$^{our}$ &Cas. & \textbf{46.4} & \textbf{64.2} & \textbf{50.4} & \textbf{40.3} & \textbf{61.8} &  \textbf{43.8}  \\
		\bottomrule[1pt]
	\end{tabular}
\end{table}

\subsection{Transfer Learning}

\textbf{VOC and CityScapes results.} We then transfer our pretrained models to VOC detection using Faster R-CNN R50-C4. We run MoCo-v2, BYOL and DenseCL with or without `Our-Crop' during the pretraining stage. As shown in Table~\ref{table:VOC-C4-city}, the improvement of our cropping strategy over MoCo-v2 and BYOL is consistent. Specifically for MoCo-v2, we get 1.1\% and 1.7\% gain over its baseline for $AP$ and ${AP_{75}}$, respectively. Our method also surpasses the ImageNet supervised baseline and the SSL method Self-EMD~\cite{self-EMD} by a large margin. 
Finally, we validate our cropping method on Cityscapes semantic segmentation. As shown in Table~\ref{table:VOC-C4-city}, using `Our-Crop' during the pretraining stage leads to consistent improvement over 3 SSL baseline methods, surpassing its supervised counterpart for all of them.

\begin{table}
	\setlength{\tabcolsep}{4.0pt}
	\centering
	\footnotesize
	\caption{Transfer learning of VOC0712 object detection using Faster-RCNN R50-C4 and CityScapes (`City') segmentation using PSANet~\cite{PSANet}. Our models were pretrained using original image and `Our-Crop' ($our$) on MS-COCO for the default 400 epochs.}
	\label{table:VOC-C4-city}
	\begin{tabular}{llllllll}
		\toprule[1pt]
		\multirow{2}{*}{Method} & \multirow{2}{*}{Data} & \multicolumn{3}{c}{VOC detection} & \multicolumn{3}{c}{City segmentation}\\
		& & AP &  AP$_{50}$ & AP$_{75}$ & mIoU & mAcc & aAcc \\
		\midrule[1pt]
		 \textcolor{lightgray}{Supervised} &  \textcolor{lightgray}{IN} & \textcolor{lightgray}{53.3} &  \textcolor{lightgray}{81.0} &  \textcolor{lightgray}{58.8} &  \textcolor{lightgray}{77.5} &  \textcolor{lightgray}{86.5} &  \textcolor{lightgray}{95.9} \\
		Self-EMD~\cite{self-EMD} & CC & 53.0 &  80.0 & 58.6 & - & - & -\\
		\midrule
		BYOL& CC & 51.7 & 80.2 & 56.4 & 77.6 & 86.6 & 95.8 \\
		BYOL$^{our}$ & CC & \textbf{52.1}  & \textbf{80.7} & \textbf{56.9} & \textbf{78.1} & \textbf{86.8} & \textbf{96.0}\\
		\midrule
		MoCo-v2 & CC & 53.7  & 80.0 & 59.5 & 76.8 & 85.7 & 95.8 \\
		MoCo-v2$^{our}$ & CC & \textbf{54.8} & \textbf{81.0} & \textbf{61.2} & \textbf{77.6} & \textbf{85.9} & \textbf{96.0} \\
		\midrule
		DenseCL & CC & 56.0 & 81.5 & 62.5 & 77.6 & 86.6 & 96.0 \\
		DenseCL$^{our}$ & CC & \textbf{57.2} & \textbf{82.2} & \textbf{63.4} & \textbf{78.6} & \textbf{86.9} & \textbf{96.2} \\
		\bottomrule[1pt]
	\end{tabular}
\end{table}

\textbf{Classification results.} Besides dense prediction tasks, we also verify the effectiveness of Our-Crop on various classification benchmarks. The results can be found in Table~\ref{table:classification}, which again clearly demonstrates the adaptability of our cropping strategy to different downstream tasks.

\begin{table}
	\setlength{\tabcolsep}{4pt}
	\caption{Transfer learning results on 7 small classification datasets, All the models are pretrained on MS-COCO dataset for 400 epochs with BYOL SSL method using original image and our cropping strategies (`\emph{our}'), respectively.}
	\label{table:classification}
	\centering
	\footnotesize
	\begin{tabular}{lllllllll}
		\toprule[1pt]
		  Method  & CUB &  Flowers & Cars & Aircraft & Indoor & Pets & DTD \\
		\midrule[1pt]
		 BYOL & 74.0 & 92.9 & 89.7 & 84.6 & 70.3 & 83.8 &  65.3\\
		 BYOL$^{our}$  & \textbf{75.4} & \textbf{94.5} & \textbf{90.4} & \textbf{85.2} & \textbf{71.0} & \textbf{84.7} & \textbf{66.2}\\
		 \bottomrule[1pt]
	\end{tabular}
\end{table}

\textbf{ImageNet pretraining.}
Now we further validate on object-centric images. We first compare Our-Crop with Selective Search~\cite{SS} (SS) under the SoCo pretraining pipeline on ImageNet. Number of boxes in SS and ours is 41 and 22 per image, respectively. As shown in Table~\ref{table:imagenet-results}, larger dataset pretraining boosts downstream performance. Although the number of crops in Our-Crop is only 22 (about half of that in SS), SoCo$^{our}$ consistently surpassed SoCo$^{ss}$. We also sample 100 images per ImageNet class to form an ImageNet subset (`IN100'), and adopt our cropping pipeline (cf. Sec.~\ref{sec:pipeline}) to generate 3 crops per image. As Table~\ref{table:imagenet-results} shows, training BYOL and MoCo-v2 with Our-Crop is significantly better than training with original image on IN100, demonstrating the flexibility of our cropping pipeline to not only non-curated datasets (\eg, MS-COCO), but to object-centric datasets (\eg, ImageNet) as well!

\begin{table}
	\setlength{\tabcolsep}{1.8pt}
	\footnotesize
	\centering
	\caption{ImageNet pretraining results. SoCo~\cite{SoCo} were pretrained with its original boxes from selective search ($ss$) or Our-Crop (`\emph{our}'). BYOL and MoCo-v2 were pretrained for 200 epochs on ImageNet subset (`IN100', with 100 sampled images per class) using Our-Crop (3 crops per image). All pretrained models were then finetuned on COCO with Mask R-CNN R50-FPN.}
	\label{table:imagenet-results}
	\begin{tabular}{lllllllll}
		\toprule[1pt]
		\multirow{2}{*}{Method} & \multirow{2}{*}{Data}  &
		\multicolumn{3}{c}{Detection} & \multicolumn{3}{c}{Segmentation}  \\

		&  & AP$^{bbox}$ & AP$^{bbox}_{50}$ & AP$^{bbox}_{75}$ & AP$^{seg}$ & AP$^{seg}_{50}$ & AP$^{seg}_{75}$\\
		\midrule[1pt]
		\textcolor{lightgray}{Supervised} &  \textcolor{lightgray}{IN} & \textcolor{lightgray}{38.9} & \textcolor{lightgray}{59.6} & \textcolor{lightgray}{42.7} & \textcolor{lightgray}{35.4} & \textcolor{lightgray}{56.5} & \textcolor{lightgray}{38.1}   \\
		SoCo$^{ss}$ & IN  & 41.7 & 62.2 & 45.7 & 37.3 & 59.1 & 39.9 \\

		SoCo$^{our}$  & IN &  \textbf{42.0} & \textbf{62.5} & \textbf{46.1} & \textbf{37.5} & \textbf{59.4} & \textbf{40.0}\\
		\midrule
		BYOL  & \multirow{2}{*}{IN100} & 37.9 & 57.4 & 41.3 & 34.3 & 54.5 & 36.7    \\
		BYOL$^{our}$ & & \textbf{38.8} & \textbf{58.5} & \textbf{42.4} & \textbf{35.2} & \textbf{55.7} & \textbf{37.5} \\
		\midrule
		MoCo-v2   & \multirow{2}{*}{IN100}  & 36.6 & 55.7 & 39.9 & 33.3 & 53.0 & 35.7  \\

		MoCo-v2$^{our}$ & & \textbf{37.8}  & \textbf{57.3} & \textbf{41.2} & \textbf{34.3} & \textbf{54.5} & \textbf{36.8}\\
		\midrule
		DenseCL   & \multirow{2}{*}{IN100}  & 37.9 & 57.4 & 41.4 & 34.5 & 54.5 & 37.0  \\

		DenseCL$^{our}$ & & \textbf{39.1} & \textbf{58.7} & \textbf{42.9} & \textbf{35.4} & \textbf{55.8} & \textbf{38.2} \\
        \bottomrule[1pt]
	\end{tabular}
\end{table}

\section{Ablations}

In this section, we will first explore various components in our cropping pipeline (Sec.~\ref{sec:ablation:efficiency}), then compare with other box generation methods in Sec.~\ref{sec:ablation-object-discovery} to verify of our hypothesis: \emph{crops that have coarse objectness scores work the best}. Besides, we do a series of hyper-parameter tuning and visualization in Sec.~\ref{sec:ablation-hyper-parameter} to show the robustness of our method.

\subsection{Efficiency and validity analysis}
\label{sec:ablation:efficiency}

\textbf{More or less training epochs.} Since there are 5 (`Our-Crop') crops per image, we explored a 5x baseline for `Image-Crop'. As shown in Table~\ref{table:long-epoch}, more training epochs will \emph{not} increase the accuracy for `Image-Crop', and that pretraining with `Our-Crop' is consistently better than `Image-Crop' with both 1x and 5x schedule. Besides, we downscale the pre-training epochs on MS-COCO from the \emph{default} 400 to 200 and 100. As shown in Table~\ref{table:less-epoch}, even with one fourth (100ep) of the default scheduler, Our-Crop is still better than Image-Crop. It is also viable for Our-Crop to adopt half scheduler since 400 and 200 epochs lead to quite similar results. 

\begin{table}
	\centering
	\footnotesize
	\setlength{\tabcolsep}{1.7pt}
	\caption{More training epochs (2000) of `Image-Crop' pretrained on COCO, and finetuned on COCO with Mask R-CNN R50-C4.}
	\label{table:long-epoch}
	\begin{tabular}{llllllll}
		\toprule[1pt]
		\multirow{2}{*}{Arch} & \multirow{2}{*}{Epochs}  & \multicolumn{3}{c}{Detection} & \multicolumn{3}{c}{Segmentation}  \\

		& & AP$^{bbox}$ & AP$^{bbox}_{50}$ & AP$^{bbox}_{75}$ & AP$^{seg}$ & AP$^{seg}_{50}$ & AP$^{seg}_{75}$\\
		\midrule[1pt]
		BYOL  & 400 & 36.9 & 56.7 & 39.4 & 32.4 & 53.5 & 34.3    \\
		BYOL  & 2000 & 36.1 & 56.3 & 38.4 & 31.7 &52.6 & 33.3   \\
		BYOL$^{our}$ & 400 &  \textbf{37.9} & \textbf{57.8} & \textbf{40.7} & \textbf{33.1} & \textbf{54.3} & \textbf{35.2}\\
		\midrule
		MoCo-v2  & 400 & 37.3 & 56.7 & 40.4 & 32.8 & 53.5 & 34.9    \\
		MoCo-v2  & 2000 & 37.4 & 56.6 & 40.3 & 32.8 & 53.3 & 34.8 \\
		MoCo-v2$^{our}$ & 400 &  \textbf{38.1} & \textbf{57.4} & \textbf{41.4} & \textbf{33.4} & \textbf{54.2} & \textbf{35.8}\\
		\bottomrule[1pt]
	\end{tabular}
\end{table}

\begin{table}
	\centering
	\footnotesize
	\setlength{\tabcolsep}{3pt}
	\caption{Less pre-training epochs of BYOL and MoCo-v2 with `Our-Crop' on COCO and finetuned with Mask R-CNN R50-FPN.}
	\label{table:less-epoch}
	\begin{tabular}{lllllllll}
	\toprule[1pt]
        \multirow{2}{*}{Crop types} & \multirow{2}{*}{Epochs}  & \multirow{2}{*}{Schedule} & \multicolumn{2}{c}{BYOL} & \multicolumn{2}{c}{MoCo-v2} \\
		 & &  & AP$^{bbox}$  & AP$^{seg}$  & AP$^{bbox}$  & AP$^{seg}$ \\
	\midrule[1pt]
		 Image-Crop & 400 & 1.0x & 38.8 & 35.0 & 38.2 & 34.7   \\
	\midrule
		\multirow{3}{*}{Our-Crop}  & 100 & 0.25x & 39.2 & 35.4 & 38.5 &  34.9   \\
		 &  200 & 0.5x & 40.1 & 36.2 & 39.2 & 35.5 \\

   &  400 & 1.0x & \textbf{40.2} & \textbf{36.4} & \textbf{39.3} & \textbf{35.6} \\
        \bottomrule[1pt]
	\end{tabular}
\end{table}

\textbf{Components of Our-Crop.} We pretrain BYOL (Table~\ref{table:COCO-FPN-stages}) with Image-Crop (`baseline'), GTpad-Crop (`s2$^{GTpad}$'), Grid-Crop (`s2$^{grid}$'), Multi-Crop (`s3$^{multi}$') and three variants of Our-Crop (`s4-s6'). Specifically, `s4' means \emph{randomly} generated boxes (pseudo-centric crops, 5 per image) for SSL learning, `s5' using \emph{randomly} initialized network weights for box filtering, and `s6' is the default Our-Crop pipeline. It is clear that `s1$^{GTpad}$' and `s2$^{grid}$' both surpassed baseline, showing that cropping is the \emph{key} (generating views from local object contents). And our cropping pipeline is effective than the pure multi-crop strategy. However, $GTpad$ (39.6\% AP$^{bbox}$) is still inferior to Our-Crop (40.2\% AP$^{bbox}$), suggesting \emph{groundtruth bounding boxes are not necessarily the most important information in SSL}. Surprisingly, we can also obtain similar results without `\emph{Stage1}' pretraining (the strategy `s5'), showing that our cropping strategy \emph{does not} rely on pretrained SSL model!

\begin{table}
	\setlength{\tabcolsep}{3.6pt}
	\footnotesize
	\centering
	\caption{Comparison of different cropping strategies illustrated in Fig.~\ref{fig:crop} and the analysis of 3 components (stages) in our cropping pipeline. We consider Image-Crop (`s0'), GTpad-Crop (`s1$^{GTpad}$'), Grid-Crop (`s2$^{grid}$'), Multi-Crop~\cite{SwAV,NIPs_revisiting} (`s3$^{multi}$') and three strategies (`s4-s6') of Our-Crop. The three stages are detailed in Sec.~\ref{sec:pipeline}, which means learning backbone weights for SCDA scores (cf. Fig.~\ref{fig:main-method}), anchors generation \& filtering, and the final SSL learning with pseudo object-centric images, respectively.}
	\label{table:COCO-FPN-stages}
	\begin{tabular}{lccccll}
	\toprule[1pt]
	 Strategy	&\emph{Stage1}  & \emph{Stage2} & \emph{Stage3} & GT box  & AP$^{bbox}$  & AP$^{seg}$\\
	\midrule[1pt]
      \multicolumn{3}{l}{\textcolor{lightgray}{\textit{different cropping}}} \\
       s0 & &  &  &   &  38.8 & 35.0 \\
    s1$^{GTpad}$ & &  & \ding{51} & \ding{51} & 39.6 & 35.9 \\
    s2$^{grid}$ & &  & \ding{51} &  & 39.0 & 35.2 \\
    s3$^{multi}$ & &  & \ding{51} &  & 39.5 & 35.7 \\

    \midrule
      \multicolumn{3}{l}{\textcolor{lightgray}{\textit{our cropping}}} \\
    s4 & &  & \ding{51} &  & 39.1 & 35.5 \\
    s5 & & \ding{51} & \ding{51}  & &  40.0 & 36.2 \\
    s6 & \ding{51} & \ding{51} & \ding{51} &  &  \textbf{40.2} & \textbf{36.4} \\
       \bottomrule[1pt]
	\end{tabular}
\end{table}

\subsection{Comparing with existing object discovery methods}
\label{sec:ablation-object-discovery}

\textbf{Quantitative comparison.} We also compare our cropping strategy with two unsupervised object discovery methods, selective search~\cite{SS} (SS) and EdgeBox~\cite{EdgeBox}. We use them to generate 5 top boxes for each image (because we have 5 crops in `Our-Crop'), and then crop them out as pseudo images. We first pretrain all models on the MS-COCO dataset. As can be seen in Table~\ref{table:other-crop_coco}, all cropping strategy (EdgeBox, SS and Our) lead to consistent improvement over the Image-Crop baseline, showing local object contents suits SSL better. Besides, our cropping strategy achieves the highest results among all of them. The same is true for VOC pretraining results (shown in Table~\ref{table:other-crop}), which clearly demonstrate the superiority of our cropping strategy over selective search and EdgeBox. Meanwhile, it is obvious that generating two views from a cropped pseudo image (SS and EdgeBox) significantly outperforms `Image-Crop'. This observation further shows that our pipeline is effective: it's better to obtain two views from a cropped pseudo image rather than from the entire scene images.

\begin{table}
	\setlength{\tabcolsep}{2pt}
	\footnotesize
	\centering
	\caption{Compare with other box generation methods (Selective Search `ss' and EdgeBox `edge') on MS-COCO object detection and segmentation. All models were SSL pretrained on MS-COCO using BYOL and MoCo-v2 for 400 epochs.}
	\label{table:other-crop_coco}
	\begin{tabular}{lllllll}
		\toprule[1pt]
		\multirow{2}{*}{Method}  & \multicolumn{3}{c}{Detection} & \multicolumn{3}{c}{Segmentation}  \\

		& AP$^{bbox}$ & AP$^{bbox}_{50}$ & AP$^{bbox}_{75}$ & AP$^{seg}$ & AP$^{seg}_{50}$ & AP$^{seg}_{75}$\\
		\midrule[1pt]
		BYOL  & 38.8 & 58.5 & 42.2 & 35.0 & 55.9 & 38.1 \\
            BYOL$^{ss}$  & 39.3 & 59.4 & 42.9 & 35.6 & 56.3 & 38.3\\
		BYOL$^{edge}$  & 39.6 & 59.8 & 43.6 & 35.9 & 56.7 & 38.5 \\
		BYOL$^{our}$   &  \textbf{40.2} & \textbf{60.4} & \textbf{43.9} & \textbf{36.4} & \textbf{57.3} & \textbf{39.0}\\
		\midrule
		MoCo   &  38.2 & 58.0 & 41.9 & 34.7 & 55.1 & 37.2 \\
            MoCo$^{ss}$ & 38.1 & 57.4 & 41.7 & 34.3 & 54.4 & 36.9 \\
		MoCo$^{edge}$  & 38.5 & 58.2 & 42.0 & 35.0 & 55.4 & 37.5 \\
		MoCo$^{our}$   &   \textbf{39.3} & \textbf{59.1} & \textbf{42.7} & \textbf{35.6} & \textbf{56.4} & \textbf{38.0}\\
		\bottomrule[1pt]
	\end{tabular}
\end{table}

\begin{table}
	\setlength{\tabcolsep}{4pt}
	\centering
	\footnotesize
	\caption{Compare with other box generation methods on VOC2007 detection and multi-label classification tasks. Models were SSL pretrained on the VOC2007 trainval set.}
	\label{table:other-crop}
	\begin{tabular}{lllllll}
		\toprule[1pt]
		Method & Crop types   & AP$_{50}^{bbox}$  &  mAP$^{l}$ \\
		\midrule[1pt]

		\multirow{4}{*}{BYOL} & Image-Crop  & 63.3 & 39.8 \\
							& Selective search & 68.0   & 36.1  \\
						&	EdgeBox & 68.1 & 30.5 \\
							& Our-Crop & \textbf{69.5} & \textbf{42.5} \\
		\midrule
		\multirow{4}{*}{MoCo-v2} & Image-Crop  & 61.8  & 26.9 \\
								& Selective search & 66.4  &  39.7\\
						&	EdgeBox	& 67.0 & 43.3 \\
								& Our-Crop & \textbf{67.2}  & \textbf{44.9} \\
		\bottomrule[1pt]
	\end{tabular}
\end{table}

\begin{figure}
    \centering
    \subfigure[MS-COCO results]
    {
    \centering
	\includegraphics[width=0.35\linewidth]{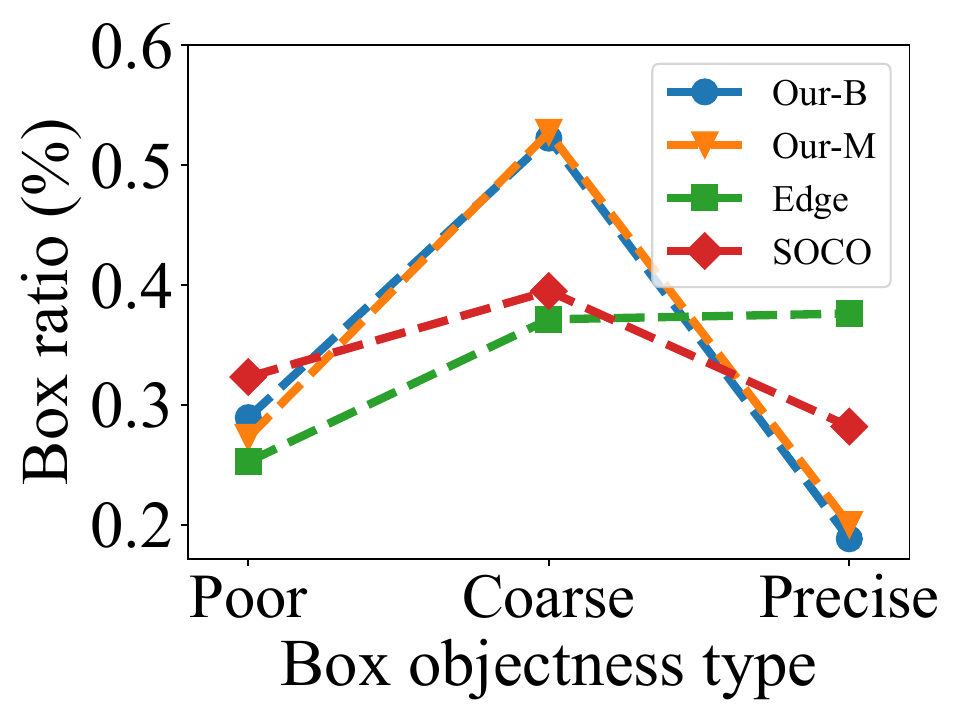}
    }
    \hspace{20pt}
    \subfigure[VOC results]
    {
    \centering
	\includegraphics[width=0.35\linewidth]{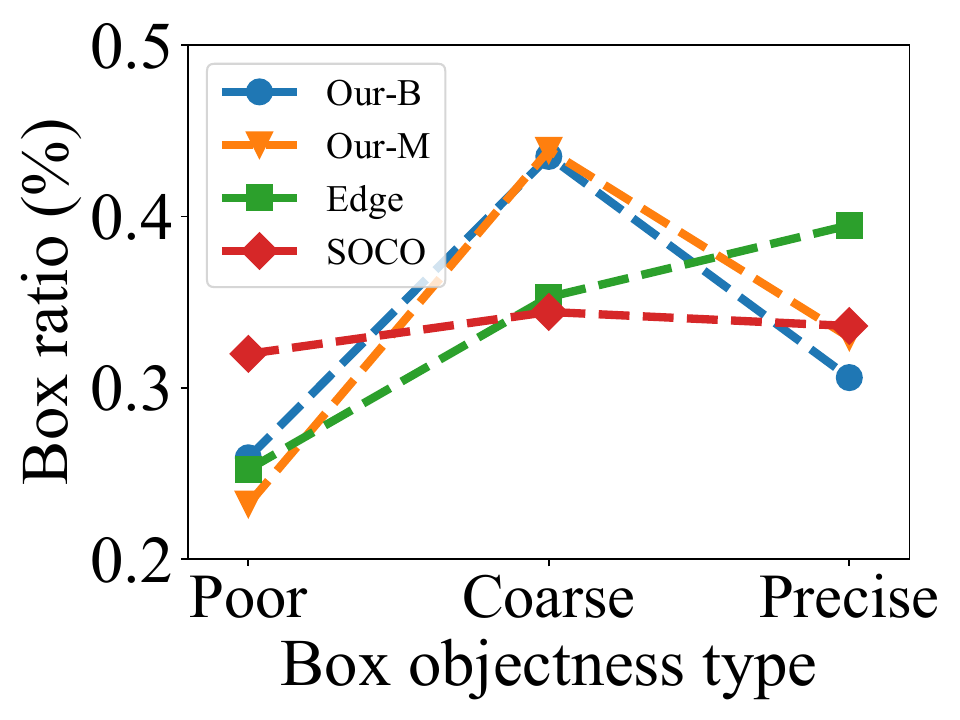}
    }    
	\caption{Objectness scores of object crops generated by our method and other object discovery methods. `Our-B' and `Our-M' means the backbone weights (used for box filtering, cf. Fig.~\ref{fig:main-method}) are obtained with BYOL and MoCo-v2 pretraining methods, respectively.}
	\label{fig:other_crop_boxtype_ratio}
\end{figure}

\textbf{Why Our-Crop is better?} We then dig deeper into the phenomenon demonstrated in Tables~\ref{table:other-crop_coco} and~\ref{table:other-crop}, that is, \emph{why our cropping strategy is better than existing object discovery methods?}. We analyze this question from the objectness aspect. Since previous findings in Table~\ref{table:crop-objectness} suggest that boxes with too high (precise crop) or too low (poor crop) objectness score are \emph{both} detrimental for SSL pretraining, we calculate the ratio of boxes that fall into each category: Poor-Crop (objectness score $<$ 20\%), Precise-Crop (objectness score $>$ 80\%) and Coarse-Crop (the remaining) for EdgeBox, Selective Search and Our-Crop. Fig.~\ref{fig:other_crop_boxtype_ratio} shows the result on MS-COCO and VOC2007, where `B' and `M' stands for crops filtered with BYOL and MoCo-v2 pretrained backbone, respectively. It is clear that existing object discovery seek highly accurate objects since their ratio of `Precise' box are both higher that our cropping strategy. These results support that our conjecture in Sec.~\ref{sec:framework} and our motivation in Sec.~\ref{sec:intro} are valid: \emph{crops with either too high or too low objectness scores} are sub-optimal, and coarse crops are \emph{generally} better for SSL.

\subsection{Hyper-parameter \& Visualization}
\label{sec:ablation-hyper-parameter}

\textbf{Number of crops.} Now we study how our single hyperparameter (number of crops per image, denoted as $N$) influences SSL quality. We perform SSL pretraining on the VOC2007 trainval set, then finetune on VOC multi-label recognition and object detection benchmarks. As shown in Table~\ref{table:TopN}, it consistently improves as $N$ grows larger. When $N=1$, our strategy performs fairly poorly, since $N=1$ means the only pseudo image contains merely a small part of an uncurated image and the global context has been lost. Bigger $N$ can guarantee enough objects of interest and indeed boost finetuning accuracy. However, it also means higher training cost after pseudo images have been cropped out. Due to constraints on our computing power, we chose $N=5$ in our experiments.

\begin{table}
	\setlength{\tabcolsep}{4pt}
	\caption{Object detection and multi-label classification results with different $N$ (number of cropped pseudo object-centric images per image) pretrained and finetuned on VOC2007.}
	\label{table:TopN}
	\centering
	\footnotesize
	\begin{tabular}{lllll}
		\toprule[1pt]
		Method & Top-N anchors & AP$_{50}^{bbox}$ &  mAP &  mAP$^l$ \\
		\midrule[1pt]
		\multirow{5}{*}{BYOL} & Image-Crop & 63.3 & 63.0 & 39.8\\
							& $N=1$ & 60.7 & 61.2 & 31.1\\
							& $N=3$ & 67.7  & 69.3 & 36.8\\
							& $N=5$ & 69.5  & 70.7 & 42.5\\
							& $N=7$ & \textbf{71.5} &\textbf{72.7} & \textbf{48.2}  \\
		\midrule
		\multirow{5}{*}{MoCo-v2} & Image-Crop & 61.8 & 62.2 & 26.9\\
								& $N=1$ & 60.1  & 60.2  & 22.6\\
								& $N=3$ & 65.6 & 68.7 & 32.4\\
								& $N=5$ & 67.2 & 69.6 & 44.9\\
							& $N=7$ & \textbf{69.3} &\textbf{72.1} & \textbf{51.2}  \\
		\bottomrule[1pt]
	\end{tabular}
\end{table}

\textbf{Augmentation sensitivity.} Since the pseudo object-centric images (the proposed Our-Crop) might already contain object of interest, we now validate whether the most important SSL data augmentation RandomResizedCrop is necessary for Our-Crop. We adopt ResNet-50 models pretrained on VOC2007 trainval with Image-Crop and Our-Crop, and separately remove the RandomResizedCrop operation for each pipeline. The VOC07 downstream object detection results can be found in Table~\ref{table:dataaug}. It is obvious that models pretrained with Our-Crop (4.5\% AP$^{bbox}_{50}$ drop) is less sensitive to that with Image-Crop (9.1\% AP$^{bbox}_{50}$ drop). Our-Crop without RandomResizedCrop even surpasses Image-Crop with this augmentation by 1.7\% AP$_{50}^{bbox}$!

\begin{table}
	\setlength{\tabcolsep}{4pt}
	\caption{VOC2007 object detection results. We pretrained BYOL on VOC2007 trainval with (w/) or without (w/o) the standard RandomResizedCrop augmentation (`RRC-aug') on original image (`Image-Crop') or the pseudo-centric images (Our-Crop).}
	\label{table:dataaug}
	\centering
	\footnotesize
	\begin{tabular}{lllll}
	\toprule[1pt]
        Method &  Crop types & RRC-aug & AP$_{50}^{bbox}$ & AP drop \\
        \midrule[1pt]
        \multirow{2}{*}{BYOL} & \multirow{2}{*}{Image-Crop} & w/ & \textbf{63.3} & (-0.0) \\
        &  & w/o & 54.2 & (-9.1) \\
        \midrule
         \multirow{2}{*}{BYOL} & \multirow{2}{*}{Our-Crop} & w/ & \textbf{69.5} & (-0.0) \\
        & & w/o & 65.0 & (-4.5) \\ 
	\bottomrule[1pt]
	\end{tabular}
\end{table}

\begin{figure}
    \centering
    \subfigure
    {
    \centering
		\includegraphics[width=0.11\linewidth]{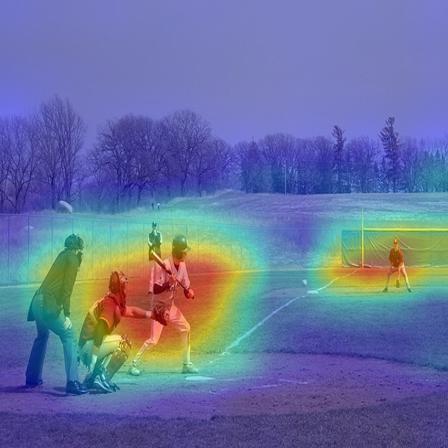}
		\includegraphics[width=0.11\linewidth]{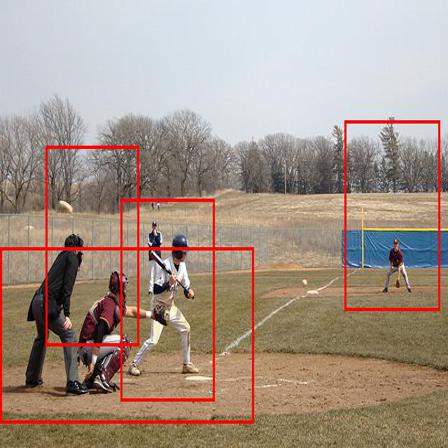}
		\includegraphics[width=0.11\linewidth]{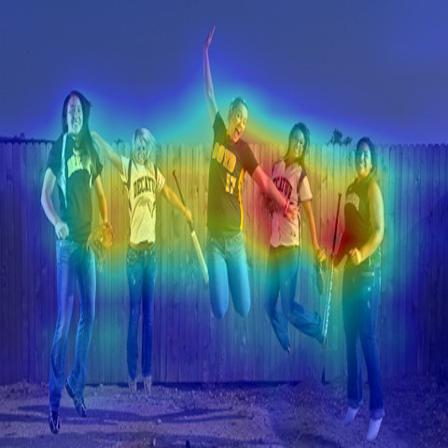}
		\includegraphics[width=0.11\linewidth]{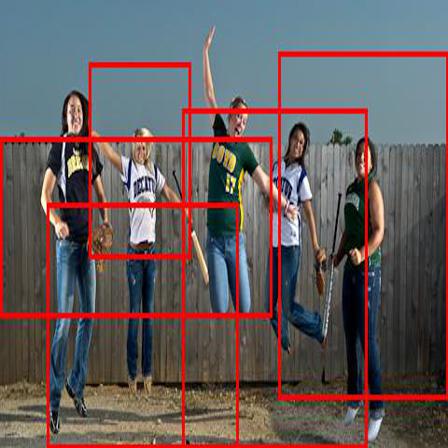}
		\includegraphics[width=0.11\linewidth]{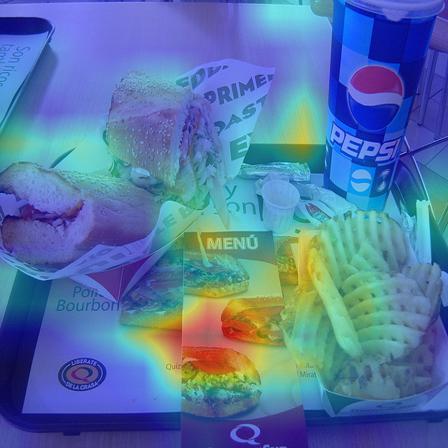}
		\includegraphics[width=0.11\linewidth]{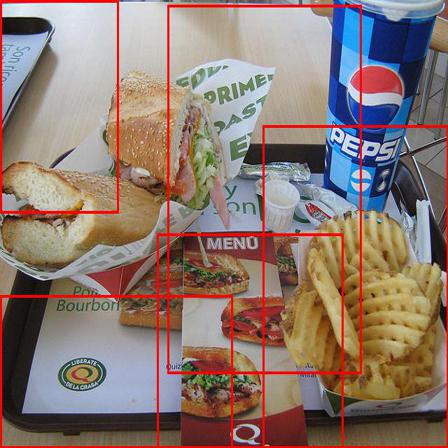}
		\includegraphics[width=0.11\linewidth]{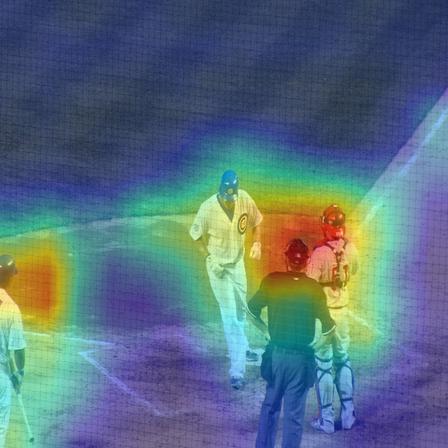}
		\includegraphics[width=0.11\linewidth]{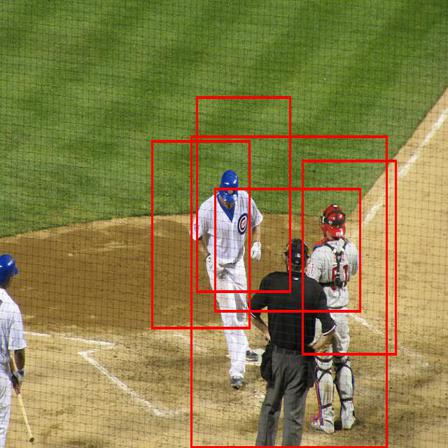}
    }

    \subfigure
    {
    \centering
		\includegraphics[width=0.11\linewidth]{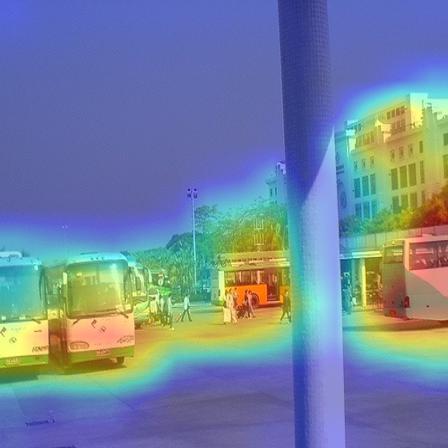}
		\includegraphics[width=0.11\linewidth]{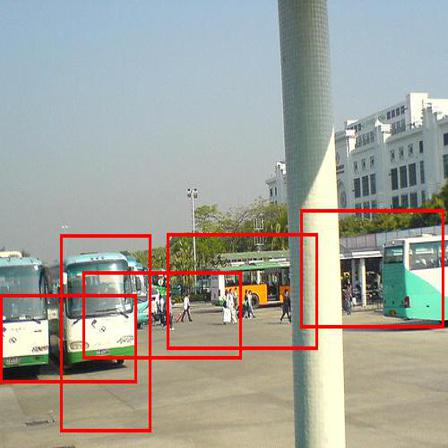}
		\includegraphics[width=0.11\linewidth]{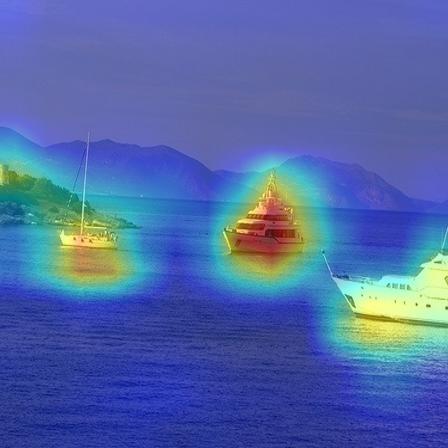}
		\includegraphics[width=0.11\linewidth]{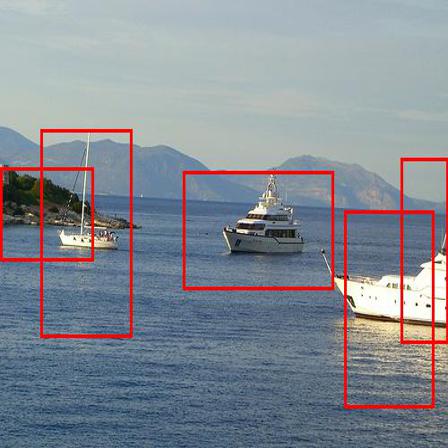}
		\includegraphics[width=0.11\linewidth]{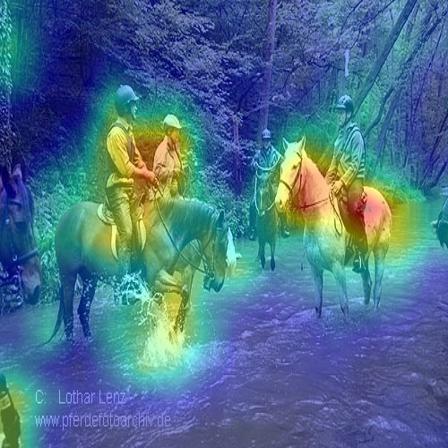}
		\includegraphics[width=0.11\linewidth]{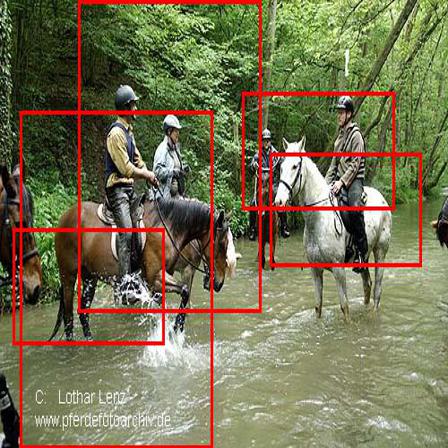}
		\includegraphics[width=0.11\linewidth]{visualize/voc/image/000328.jpg}
		\includegraphics[width=0.11\linewidth]{visualize/voc/box/000328.jpg}
    }
	\caption{Visualization of our objectness score map and the cropped boxes on COCO (1st row) and VOC (2nd row) images. The backbone that generates the objectness scores were pretrained using MoCo-v2 on COCO \& VOC `Image-Crop', respectively. Best viewed in color.}
	\label{fig:visualization}
\end{figure}

\textbf{Visualization.} At last, we randomly sampled a few images from the VOC2007 trainval set and the MS-COCO train2017 set, then calculated their normalized score maps $\hat{A}$ (cf. Fig~\ref{fig:main-method}). The left image of each pair is the normalized map $\hat{A}$, while the right one shows the final generated coarse cropping boxes with top five scores. Note that the score maps are color coded (red and blue are high and low scores, respectively). As can be seen in Fig~\ref{fig:visualization}, our normalized score map effectively captures objects, and filters out unnecessary background. These attention maps again demonstrate the power of current SSL methods: an SSL pretrained model's weights are already indicative of objects~\cite{Tobias}.

\begin{figure}
    \centering
    \subfigure[Object detection task]
    {
    \centering
	\includegraphics[width=0.35\linewidth]{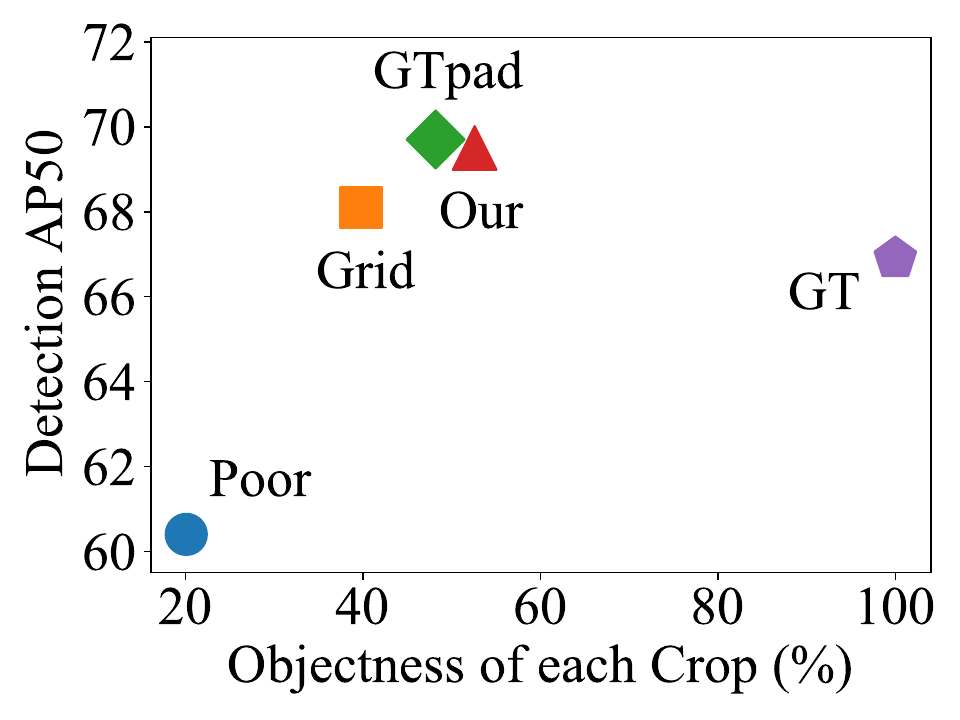}
    }
    \hspace{25pt}
    \subfigure[Multi-label recognition task]
    {
    \centering
	\includegraphics[width=0.35\linewidth]{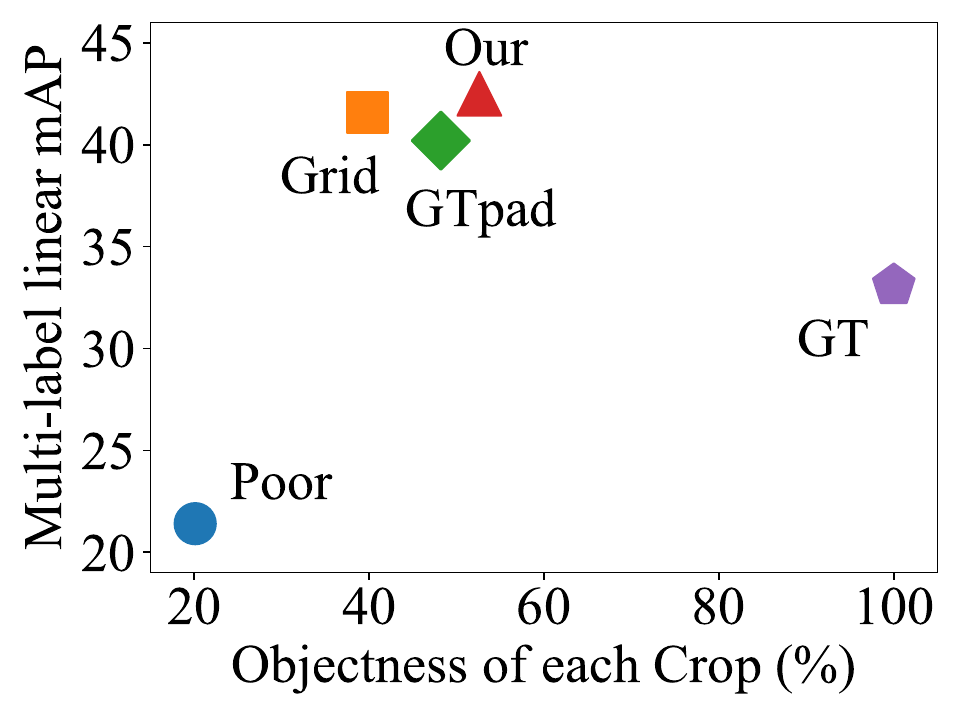}
    }    
	\caption{Results of VOC07 downstream tasks with different types of pretraining crops and their (averaged) objectness scores. Models are SSL pretrained with BYOL on VOC07 trainval.}
	\label{fig:objness-mAP}
\end{figure}

We also plot the relation between objectness of each cropping strategy and its corresponding results in Fig.~\ref{fig:objness-mAP} (also shown in Table~\ref{table:crop-objectness}), which clearly supports our motivation: \emph{crops with coarse objects work the best}.

\section{Conclusions}

In this paper, we argued that successfully cropping coarse objects out benefits self-supervised learning (SSL) with uncurated scene images. Based on this finding, we designed a novel and efficient crop generation strategy, which utilizes the unsupervised model weights and anchor generator to obtain a few coarse crops with varying shapes and sizes. The crops (pseudo object-centric images) are then fed into object-centric SSL methods. Experiments on detection and segmentation benchmarks have clearly verified the effectiveness of our pipeline over other SSL methods on uncurated scene images. We have also carefully designed a series of ablations to verify the efficiency and validity of our cropping strategy from both empirical and theoretical aspects.

As for the limitations, it remains unclear what's the best objectness score (a certain number) of the pseudo-centric images that benefits \emph{SSL pretraining}. We argued that mid-level objectness works better since it contains both objects and contexts. There might be other statistical metrics besides objectness that could also evaluate the quality of each cropping strategy. In the future, we will fully explore these questions that are helpful for various downstream tasks.

{\small
\bibliographystyle{ieee_fullname}
\bibliography{egbib}
}

\end{document}